\documentclass{article}
\usepackage{spconf,amsmath,graphicx}

\usepackage{amsmath,amsfonts,bm}

\def\eqref#1{equation~\ref{#1}}

\def\1{\bm{1}}

\def\ve{{\bm{e}}}
\def\vf{{\bm{f}}}

\def\vh{{\bm{h}}}

\def\vp{{\bm{p}}}
\def\vq{{\bm{q}}}

\def\vu{{\bm{u}}}
\def\vv{{\bm{v}}}
\def\vw{{\bm{w}}}
\def\vx{{\bm{x}}}
\def\vy{{\bm{y}}}
\def\vz{{\bm{z}}}

\def\evf{{f}}

\def\evz{{z}}

\def\mA{{\bm{A}}}
\def\mB{{\bm{B}}}

\def\mU{{\bm{U}}}
\def\mV{{\bm{V}}}
\def\mW{{\bm{W}}}
\def\mX{{\bm{X}}}

\DeclareMathAlphabet{\mathsfit}{\encodingdefault}{\sfdefault}{m}{sl}
\SetMathAlphabet{\mathsfit}{bold}{\encodingdefault}{\sfdefault}{bx}{n}

\def\gO{{\mathcal{O}}}

\def\sR{{\mathbb{R}}}

\newcommand{\vectorize}{\mathrm{vec}}

\DeclareMathOperator*{\argmin}{arg\,min}

\def\eg{\emph{e.g.~}}

\def\etc{\emph{etc.~}}

\def\ourlayer{JCF~}
\newcommand*\rfrac[2]{{}^{#1}\!/_{#2}}

\title{Efficient Codebook and Factorization for Second Order Representation Learning}

\name{Pierre Jacob$^1$ \qquad David Picard$^1$ \qquad Aymeric Histace$^1$ \qquad Edouard Klein$^2$}
\address{$^1$ETIS UMR 8051, Universit{\'e} Paris Seine, UCP, ENSEA, CNRS, F-95000, Cergy, France \\
         $^2$C3N, P\^{o}le Judiciaire de la Gendarmerie Nationale, 5 boulevard de l'Hautil, 95000 Cergy, France\\
          \small \{pierre.jacob, picard, aymeric.histace\}@ensea.fr}

\begin{document}
\ninept
\maketitle

\begin{abstract}
    Learning rich and compact representations is an open topic in many fields such as object recognition or image retrieval.
    Deep neural networks have made a major breakthrough during the last few years for these tasks but their representations are not necessary as rich as needed nor as compact as expected.
    To build richer representations, high order statistics have been exploited and have shown excellent performances, but they produce higher dimensional features.
    While this drawback has been partially addressed with factorization schemes, the original compactness of first order models has never been retrieved, or at the cost of a strong performance decrease.
    Our method, by jointly integrating codebook strategy to factorization scheme, is able to produce compact representations while keeping the second order performances with few additional parameters.
    This formulation leads to state-of-the-art results on three image retrieval datasets.
\end{abstract}


\section{Introduction}
    Learning rich and compact representations is an open topic in many fields such as object recognition \cite{Szegedy_2015_CVPR} or image retrieval \cite{Opitz_2017_ICCV, Carvalho_2018_SIGIR}.
    Recently, representations that compute first order statistics over input data have been outperformed by improved models that compute higher order statistics \cite{Perronnin_2010_ECCV, Picard_2011_ICIP, Picard_2016_ICIP, Jacob_2018_ICIP}.
    This strategy generates richer representations and are the state-of-the-art methods on fine grained visual classification tasks \cite{Lin_2015_ICCV}.
    
    However, even if the increase in performances is unquestionable, second order models suffer from a collection of drawbacks: quadratically increasing dimensionality, costly dimensionality reduction, difficulty to be trained, lack a proper adapted pooling.
    
    The two main downsides, namely the high dimensional output representations and the sub-efficient pooling scheme, have been widely studied over the last decade.
    On the one hand, the dimensionality issue has been studied through factorization scheme, either representation oriented \cite{Gao_2016_CVPR, Kim_2017_ICLR} or task oriented \cite{Kong_2017_CVPR}.
    While these factorization schemes are efficient in term of computation cost and number of parameters, the intermediate representation is still very large (typically 10k dimensions) and hinders the training process, while using lower dimension greatly deteriorates performances.
    
    On the other hand, it is well-known that global average pooling schemes aggregate unrelated features.
    This problem has been tackled by the use of codebooks (\emph{e.g.}, VLAD \cite{Arandjelovic_2013_CVPR} and Fisher Vectors \cite{Perronnin_2010_ECCV}) and extended to be end-to-end trainable \cite{Arandjelovic_2016_CVPR, Tang_2016_Arxiv}.
    However, using a codebook on second-order features leads to an unreasonably large model, since the already large feature has to be duplicated for each entry of the codebook.
    This is the case for example in MFAFVNet \cite{Li_2017_ICCV_MFAFVNet} for which the second order layer alone (\textit{i.e.}, without the CNN part) costs as much as an entire ResNet50.
    
    In this paper, we tackle the intermediate representation cost and the lack of proper pooling shortcomings by exploring joint factorization and codebook strategies.
    Our main results are the following:
    \begin{itemize}
        \item[-] We first show that state-of-the-art factorization schemes can be improved by the use of a codebook pooling, albeit at a prohibitive cost.
        \item[-] We then propose our main contribution, a joint codebook and factorization scheme that achieves similar results at a much reduced cost.
    \end{itemize}
    Since our approach focuses on representation learning and is task agnostic, we validate it in a retrieval context on several image datasets to show the relevance of the learned representations.
    We show our model achieves competitive results on these datasets at a very reasonable cost.
    
    The remaining of this paper is organized as follows: in the next section, we present the related work on second order pooling, factorization schemes and codebook strategies.
    In section \ref{sec:method_overview}, we present our factorization with the codebook strategy and how we improve its integration.
    In section \ref{sec:ablation}, we show an ablation study on the Stanford Online Products dataset \cite{Song_2016_CVPR}.
    Finally, we compare our approach to the state-of-the-art methods on three image retrieval datasets (Stanford Online Products, CUB-200-2001, Cars-196).

\section{Related work}
    
    In this section, we focus on second-order information (sections \ref{sec:rw_so_pooling} \ref{sec:rw_facto}) and on codebook strategies (section \ref{sec:rw_codebook_strategies}).

    \subsection{Second-Order Pooling}\label{sec:rw_so_pooling}
        In this section, we briefly review end-to-end trainable Bilinear pooling (BP) \cite{Lin_2015_ICCV}.
        This method extracts representations from the same image with two CNNs and computes the cross-covariance as representation.
        This representation outperforms its first-order version and other second-order representations such as Fisher Vectors \cite{Perronnin_2010_ECCV} once the global architecture is fine-tuned.
        Most of recent works on bilinear pooling only focus on computing covariance of the extracted features with a single CNN, that is :
        \begin{equation}
            \vy = \sum_{i} \vx_i \vx_i^T = \mX \mX^T \in \sR^{d \times d}
        \end{equation}
        where $\mX \in \sR^{d \times hw}$ is the matrix of the $h \times w$ extracted $d$-dimensional CNN features.
        Another formulation is the vectorized version of $\vy$ obtained by computing the Kronecker product ($\otimes$) of $\vx_i$ with itself:
        \begin{equation}\label{eq:full_bp_fomulation}
            \vy = \sum_{i} \vx_i \otimes \vx_i = \vectorize({\mX \mX^T}) \in \sR^{d^2}
        \end{equation}
        Due to the very high dimension of the above representation that is quadratic in the feature dimension, factorization schemes are mandatory.
        
    \subsection{Factorization schemes}\label{sec:rw_facto}
        Recent works on bilinear pooling proposed factorization schemes with two objectives: avoiding the direct computation of second order features and reducing the high dimensionality output representation.
        One of the main end-to-end trainable factorization is based on Tensor Sketch (CBP-TS) \cite{Gao_2016_CVPR} which tackles the high dimensionality of second-order features using sketching functions.
        Their formulation allows to keep less than 4\% of the components with nearly no loss in performances compared to the uncompressed model.
        
        This rank-one factorization has been generalized to multi-rank by taking advantage of the SVM formulation to jointly train the network and the classifier \cite{Kong_2017_CVPR}.
        Even if the second-order features are never directly computed, their factorization is limited to the SVM formulation and cannot be used for other tasks.
        Another task agnostic extensions are \emph{e.g.}, FBN \cite{Li_2017_ICCV_FBN} which also integrates the first order into the representation and HPBP \cite{Kim_2017_ICLR} which improves the factorization with attention model and non-linearity and applies it to visual question answering.
        Grassmann BP \cite{Wei_2018_ECCV} also improves second-order pooling by dealing with the "burstiness" of features which may be predominant in high order representations by using Grassmann manifolds and providing an indirect computation of the representation.
        However, their method relies on Singular Value Decomposition (SVD) and they need to greatly reduce the input feature dimension due to the SVD computation complexity which is cubic in the feature dimension.
        
        For image retrieval tasks, producing very compact representation is mandatory to tackle the indexing of very large datasets.
        \emph{E.g.}, current state-of-the-art method on the CUB dataset \cite{CUB_200_2011} named HTL \cite{Wei_2018_ECCV} uses only 512 dimensions for the representation.
        Thus, all of the aforementioned methods have representations that are still too large to compete in this category.
        In this work, we start from a rank-one factorization detailed in section \ref{sec:method_facto} which is extended by the introduction of a codebook strategy that allows smaller representation dimension, improves performances and makes them competitive to state-of-the-art methods in image retrieval.
        
    \subsection{Codebook strategies}\label{sec:rw_codebook_strategies}
        An acknowledged drawback of pooling methods is that they pool unrelated features that may decrease performances. To cope with this observation, codebook strategies (\emph{e.g.}, Bag of Words) have been proposed and greatly improved performances by pooling only features that belong to the same codeword.
        
        In the case of second order information, the first representations that take advantage of codebook strategies are VLAT \cite{Picard_2011_ICIP, Picard_2013_CVIU} and Fisher Vectors \cite{Perronnin_2010_ECCV}.
        While in VLAT the high-dimensionality is handled by PCA on local features and intra-projections, Fisher Vectors (FVs) replace the hard assignment by a Gaussian Mixture Model (GMM) and supposes that covariance matrices to be diagonal which leads to smaller representations.
        However, FV ignores cross-dimension correlations.
        Strategies like STA \cite{Picard_2016_ICIP, Jacob_2018_ICIP} extends the VLAT representation by computing cross correlation matrices of nearby features to integrate spatial information and takes advantage of a codebook strategy to avoid the computation of unrelated features.
        However, as the dimensionality is both quadratic in the codebook size and the feature dimension, factorization scheme is mandatory.
        In the case of ISTA \cite{Jacob_2018_ICIP}, the proposed dimensionality reduction only allows to reduce the dimensionality to around 20k dimensions, which is twice higher than standard second-order pooling factorization.
        
        In end-to-end trainable architectures, FisherNet \cite{Tang_2016_Arxiv} extends the FVs and outperforms non-trainable FV approaches but nonetheless has the high output dimension of the original FV.
        MFA-FV network \cite{Li_2017_ICCV_MFAFVNet}, which extends MFA-FV of \cite{Dixit_2016_NIPS}, generates an efficient representation of non-linear manifolds with a small latent space and is trainable in an end-to-end way.
        The main drawbacks of their method is the direct computation of second-order features for each codeword (computation cost), the raw projection of this covariance matrix into the latent space for each codeword (computation cost and number of parameters), and finally the representation dimension.
        In the original paper, the proposed representation reaches 500k dimensions, which is prohibitive for image retrieval as it may require more memory than whole images.
        
        To our knowledge, no efficient factorization combined with codebook strategy has been proposed to exploit the richer representation of second order features combined with the codebook strategy.
        Our propositions combine the best of both worlds by providing a joint codebook and factorization optimization scheme with a similar number of parameters and computation cost to that of methods without codebook strategies.

\section{Method overview}\label{sec:method_overview}
    After a presentation of the initial factorization (section \ref{sec:method_facto}), we first propose an extension to a codebook strategy (section \ref{sec:method_codebook_strategy}) and show the limitations of this architecture in terms of computation cost, low-rank approximation, number of parameters, \etc
    Finally, we present our shared projectors strategy (section \ref{sec:method_sharing_projectors}) which leads to a joint codebook and factorization optimization.

    \subsection{Initial factorization scheme}\label{sec:method_facto}
        In this section, we present the factorization of the projection matrix and highlight the advantages and limitations of this scheme.
        Using the same notation as in section \ref{sec:rw_so_pooling}, we want to find the optimal linear projection matrix $\mW \in \sR^{d^2 \times D}$ to build the output feature $\vz(\vx) \in \sR^D$.
        These output features are then pooled to build the output representation $\vz$:
        \begin{equation}
            \vz = \sum_{\vx} \vz(\vx) = \sum_{\vx} \mW^T (\vx \otimes \vx)
        \end{equation}
        In the rest of the paper, we use the notation $\evz_i$ that refers to the $i$-th dimension of the output representation $\vz$ and $\evz_i(\vx)$ the $i$-th dimension of the output feature $\vz(\vx)$, that is:
        \begin{equation}\label{eq:proj_so_feature}
            \evz_i = \sum_{\vx} \evz_i(\vx) = \sum_{\vx} \vw_i^T(\vx \otimes \vx) = \sum_{\vx} \left<\vw_i~;~\vx \otimes \vx\right>
        \end{equation}
        where $\vw_i \in \sR^{d^2}$ is a column of $\mW$.
        Due to the large number of parameters induced by this projection matrix, we enforce the rank one decomposition $\vw_i=\vu_i \otimes \vv_i$ where $(\vu, \vv) \in (\sR^d)^2$ for all projectors of $\mW$.
        $\evz_i(\vx)$ from Eq.\! (\ref{eq:proj_so_feature}) becomes:
        \begin{equation}\label{eq:facto_so_feature}
            \evz_i(\vx) = \left<\vu_i~;~\vx\right>\left<\vv_i~;~\vx\right>
        \end{equation}
        This factorization is efficient in term of parameters as it needs only $2dD$ parameters instead of $d^2D$ for the full projection matrix.
        However, even if this rank one decomposition allows efficient dimension reduction, it is not enough to keep all the richness of the second-order statistics due to the pooling of unrelated features.
        Consequently, we extend the second-order feature to a codebook strategy.
    
    \subsection{Codebook strategy}\label{sec:method_codebook_strategy}
        To avoid destructive averaging, we want to pool only similar features which belong to the same codeword.
        This codebook pooling is interesting because each projection to a sub-space should have only similar features.
        Thus, they lie on a simpler manifold and they could be encoded with fewer dimensions.
        For a codebook size of $N$, we compute an assignment function $\vh(\cdot) \in \sR^N$.
        This function could be a hard assignment (\eg\!, the $\argmin{}$ over distance to each cluster) or a soft assignment (\eg\!, the softmax).
        Thus, output feature $\evz_i(\vx)$ becomes:
        \begin{equation}
            \evz_i(\vx) = \left<\vw_i~;~\vh(\vx) \otimes \vx \otimes \vh(\vx) \otimes \vx\right>
        \end{equation}
        Remark that now $\mW \in \sR^{N^2d^2 \times D}$ and $\vw_i \in \sR^{N^2d^2}$.
        Here, we duplicate $\vh(\vx)$ to keep the generalization of bilinear pooling (two codebooks can be learned, one per modality) or for STA based strategies (two nearby features may belong to different codewords).
        As in equation \ref{eq:facto_so_feature}, we enforce the rank one decomposition of $\vw_i = \vp_i \otimes \vq_i$ where $(\vp_i, \vq_i) \in (\sR^{Nd})^2$ to split the modalities. This first factorization leads to the following output feature $\evz_i(\vx)$:
        \begin{equation}\label{eq:our_first_facto}
            \evz_i(\vx) = \left<\vp_i~;~\vh(\vx) \otimes \vx \right> \left<\vq_i~;~\vh(\vx) \otimes \vx\right>
        \end{equation}
        However, this representation is still too large to be computed directly.
        Then, we enforce two supplementary factorizations:
        \begin{equation}
            \begin{array}{ll}
                \vp_i & = \sum_j \ve^{(j)} \otimes \vu_{i,j} \\
                \vq_i & = \sum_j \ve^{(j)} \otimes \vv_{i,j}
            \end{array}
        \end{equation}
        where $\ve^{(j)} \in \sR^N$ is the $j$-th vector from the natural basis of $\sR^N$ and $(\vu_{i,j}, \vv_{i,j}) \in (\sR^d)^2$. 
        The decompositions of $\vp_i$ and $\vq_i$ play the same roles as intra-projection in VLAD \cite{Delhumeau_2013_ACM}.
        Indeed, if we consider $\vh(\cdot)$ as a hard assignment function, the only computed projection is the one assigned to the corresponding codewords.
        Thus, this model learns a projection matrix for each codebook entry.
        
        Furthermore, by exploiting the same property used in Eq.\! (\ref{eq:our_first_facto}), the following equation can be compacted such as:
        \begin{equation}\label{eq:our_second_facto_simplified}
             z_i(\vx) =  \left(\vh(\vx)^T \mU_i^T \vx \right) \left( \vh(\vx)^T \mV_i^T \vx \right)
        \end{equation}
        where $\mU_i \in \sR^{d \times N}$ and $\mV_i \in \sR^{d \times N}$ are the matrices concatenating the projections of all entries of the codebook for the $i$-th output dimension. We call it \emph{Joint Codebook and Factorization}, \ourlayer\!-N.
        
        This representation has multiple advantages: First, it computes second order features that leads to better performances compared to its first order counterpart.
        Second, our first factorization provides an efficient alternative in terms of number of parameters and computation despite the decreasing performances when it reaches small representation dimensions.
        This downside is addressed by the codebook strategy.
        It allows the pooling of only related features while their projections to a sub-space is more compressible.
        However, even if this codebook strategy improves the performances, the number of parameters is in $\gO(dDN)$
        As such, using large codebook may become intractable.
        In the next section, we extend this scheme by sharing a set of projectors and enhance the decompositions of $\vp_i$ and $\vq_i$.
        
    \subsection{Sharing projectors}\label{sec:method_sharing_projectors}
        In the previous section, one projector is learned to map all features that belong to a given codebook entry for each entry of the codebook.
        The proposed idea is, instead of using a one-to-one correspondence, we learn a set of projectors that is shared across the codebook.
        The reasoning behind is that projectors from different codebook entries are unlikely to be all orthogonal.
        By doing such hypothesis (\emph{i.e.}, the vector space spaned by the combination of all the projection matrices has a lower dimension than the codebook itself), we can have smaller models with nearly no loss in performances.
        To check this hypothesis, we extend the proposed factorization from section \ref{sec:method_codebook_strategy}. We want to generate $\mU_i$ from $\{\widetilde{\mU}_i\}_{i \in \{1, ..., R\}}$ and $\mV_i$ from $\{\widetilde{\mV}_i\}_{i \in \{1, ..., R\}}$ where $R$ is the number of projections in the set. Then the two new enforced factorization of $\vp_i$ and $\vq_i$ are:
        \begin{equation}
            \begin{array}{ll}
                \vp_i(\vx) & = \sum_r \evf_{p,r}\Big(\vh(\vx)\Big) \ve^{(r)} \otimes \widetilde{\vu}_{i,r} ~~ \text{and}\\
                \vq_i(\vx) & = \sum_r \evf_{q,r}\Big(\vh(\vx)\Big) \ve^{(r)} \otimes \widetilde{\vv}_{i,r}
            \end{array}
        \end{equation}
        where $\vf_{p} ~ \text{and} ~ \vf_{q}$ are two functions from $\sR^N ~ \text{to} ~ \sR^R$  that transform the codebook assignment into a set of coefficient which generate their respective projection matrices.
        Similarly to Eq.\! (\ref{eq:our_second_facto_simplified}), we have:
        \begin{equation}\label{eq:nearly_final_eq}
            z_i(\vx) =  \left(\vf_p\Big(\vh(\vx)\Big)^T \widetilde{\mU}_i^T \vx \right) \left( \vf_q\Big(\vh(\vx)\Big)^T \widetilde{\mV}_i^T \vx \right)
        \end{equation}
        In this paper, we only study the case of a linear projection:
        \begin{equation}\label{eq:truly_final_eq}
            z_i(\vx) = \left(\vh(\vx)^T \mA \widetilde{\mU}_i^T \vx \right) \left(\vh(\vx)^T \mB \widetilde{\mV}_i^T \vx \right)
        \end{equation}
        where $(\mA, \mB) \in (\sR^{N \times R})^2$.
        Eq.\! (\ref{eq:truly_final_eq}) is more efficient in terms of parameters than Eq.\! (\ref{eq:our_second_facto_simplified}) as it requires $\rfrac{R}{N}$ times lesser parameters and computation. We call this approach \ourlayer\!-N-R.
        In section \ref{sec:ablation}, we provide an ablation study of the proposed method, comparing Eq.\! (\ref{eq:our_second_facto_simplified}) and Eq.\! (\ref{eq:truly_final_eq}), demonstrating that learning recombination is both efficient and performing.

    \subsection{Implementation details} \label{sec:method_implementation}
        We build our model over pre-trained backbone network such as VGG16 \cite{Simonyan_2014_ILSVRC} (on CUB and CARS datasets) and ResNet50 \cite{He_2016_CVPR} (on Stanford Online Products).
        In both case, the features are reduced to 256d and $\ell_2$-normalized.
        The assignment function $\vh$ is the softmax over cosine similarity between the features and the codebook.
        In metric, we use Recall@K which takes the value 1 if there is at least one element from the same instance in the top-K results else 0 and averages these scores over the test set.
        Images are resized to 224x224 and we do not use data augmentation.
        We use SGD with a learning rate of $10^{-5}$, a batch of 64 images, the N-pair triplet loss \cite{Sohn_2016_NIPS} with the margin set to $0.1$ and $N=2$.
        We also use semi-hard mining for the final comparison to the state-of-the-art.

\section{Ablation studies}\label{sec:ablation}
    \subsection{Bilinear pooling and codebook strategy}\label{sec:ablation_codebook}
        \begin{table*}[t]
            \centering
            \begin{tabular}{|c|c||c|c||c||c||c|c|c||c|c|c|c|}\hline
             Method & \emph{Baseline} & \multicolumn{2}{c||}{BP} & \multicolumn{9}{c|}{\ourlayer\!-N-R} \\\hline
             N & - & - & 4 & - & 4 & \multicolumn{3}{c||}{16} & \multicolumn{4}{c|}{32} \\\hline
             R & - & - & - & - & - & 4 & 8 & 16 & 4 & 8 & 16 & 32 \\\hline
             Parameters & 1M & 34M & 135M & 0.8M & 1.6M & 1.6M & 2.6M & 4.7M & 1.6M & 2.6M & 4.7M & 8.9M \\\hline
             R@1 & 63.8 & \underline{65.9} & 67.1 & 65.0 & 65.5 & \textbf{68.2} & \textbf{68.3} & \textbf{69.8} & \textbf{68.1} & \textbf{69.4} & \textbf{69.7} & \textbf{70.6} \\\hline
            \end{tabular}
            \caption{Comparison of codebook strategy in terms of parameters and performances between the \emph{Baseline}, BP and our \ourlayer\! for different codebook size (N) and low-rank approximation (R) on Stanford Online Products.}
            \label{tab:codebook_and_bp}
        \end{table*}
        
        In this section, we demonstrate both the relevance of second-order information for retrieval tasks and the influence of the codebook on our method.
        We report recall@1 on Stanford Online Products in Table \ref{tab:codebook_and_bp} for the different configuration detailed below.
        
        First, as a reference, we train a \emph{Baseline} network, \emph{i.e.}, which consists in the average of the features reduced to 512 dimensions (first order model).
        Then we re-implement BP and extend it naively to a codebook strategy. The objective is to demonstrate that such strategy performs well, but at an intractable cost.
        Results are reported in the left part of Table \ref{tab:codebook_and_bp}.
        This experiment confirms the interest of second-order information in image retrieval with a improvement of 2\% over the baseline, while using a 512 dimension representation.
        Furthermore, using a codebook strategy with few codewords enhances bilinear pooling by 1\% more.
        However, the number of parameters becomes intractable for codebook of size greater than 4: this naive strategy requires 270M parameters to extend this model to a codebook with a size of 8.
        
        Using the factorization from Eq.\! (\ref{eq:our_second_facto_simplified}) greatly reduces the required number of parameters and allows the exploration of larger codebook.
        In the case of the factorization alone, the small representation dimension leads to poor performances and are only slightly retrieved using a codebook.
        On the opposite, our factorization which exploits both the larger codebook and the low-rank approximation is able to reach higher performances (+4\% between BP and \ourlayer\!-32-32) with nearly four times less parameters.
    
    \subsection{Sharing projections}\label{sec:ablation_sharing}
        In this part, we study the impact of the sharing projection. We use the same training procedure as in the previous section.
        For each codebook size, we train architecture with a different number of projections, allowing to compare architectures without the sharing process to architectures with greater codebook size but with the same number of parameters by sharing projectors.
        Results are reported in the right part of Table \ref{tab:codebook_and_bp}.
        Sharing projectors leads to smaller models with few loss in performances, and using richer codebooks allows more compression with superior results.
        In the next section, we compare our best model \ourlayer\!-32-32 and its shared version with four times less parameters \ourlayer\!-32-8 against state-of-the-art methods.
    
\section{Comparison to the state-of-the-art}\label{sec:results}
    In this section, we compare our method to the state-of-the-art on 3 retrieval datasets: Stanford Online Products \cite{Song_2016_CVPR}, CUB-200-2011 \cite{CUB_200_2011} and Cars-196 \cite{CARS_196}.
    For Stanford Online Products and CUB-200-2011, we use the same train/test split as \cite{Song_2016_CVPR}.
    For Cars-196, we use the same as \cite{Opitz_2017_ICCV}.
    We report the standard recall@K with $K \in \{1, 10, 100, 1000\}$ for Stanford Online Products and with $K \in \{1, 2, 4, 8\}$ for the other two.
    We implement the codebook factorization from Eq.\! (\ref{eq:our_second_facto_simplified}) with a codebook size of 32 (denoted \ourlayer\!-32).
    While \ourlayer\!-32 outperforms state-of-the-art methods on the three dataset, our low-rank approximation \ourlayer-32-8, which cost 4 times less also leads to state-of-the-art performances on two of them with a loss between 1-2\% consistent with our ablation studies.
    In the case of Cars-196 however, the performances are much more lower than the full model.
    We argue that the variety introduced by the colors, the shapes, \etc in cars requires more projections to be estimated, as it is observed for the full model.
    
    \begin{table}[!t]
            \begin{center}
                \label{tab:SOP}
                \caption{Comparison with the state-of-the-art on Stanford Online Products dataset. \textbf{bold} scores are the current state-of-the-art and \underline{underlined} are the second ones.}
                \begin{tabular}{|l|cccc|}\hline
                    r@ & 1 & 10 & 100 & 1000 \\\hline
                    LiftedStruct \cite{Song_2016_CVPR} & 62.1 & 79.8 & 91.3 & 97.4 \\
                    Binomial deviance \cite{Ustinova_2016_NIPS} & 65.5 & 82.3 & 92.3 & 97.6 \\
                    N-pair loss \cite{Sohn_2016_NIPS} & 67.7 & 83.8 & 93.0 & 97.8 \\
                    HDC \cite{Yuan_2017_ICCV} & 69.5 & 84.4 & 92.8 & 97.7 \\
                    Margin \cite{Wu_2017_ICCV} & 72.7 & 86.2 & 93.8 & 98.0 \\
                    BIER \cite{Opitz_2017_ICCV} & 72.7 & 86.5 & 94.0 & 98.0 \\
                    Proxy-NCA \cite{Movshovitz-Attias_2017_ICCV} & 73.7 & - & - & - \\
                    HTL \cite{Ge_2018_ECCV} & 74.8 & 88.3 & 94.8 & 98.4 \\
                    \hline
                    \ourlayer-32 & \textbf{77.4} & \underline{89.9} & \textbf{95.8} & \underline{98.6} \\
                    \ourlayer-32-8 & \underline{76.6} & \textbf{90.0} & \textbf{95.8} & \textbf{98.7} \\
                    \hline
                \end{tabular}
            \end{center}
        \end{table}
    
    \setlength{\tabcolsep}{2.75pt}
    \begin{table}[t]
        \begin{center}
            \caption{Comparison with the state-of-the-art on CUB-200-2011 and Cars-196 datasets. \textbf{bold} scores are the current state-of-the-art and \underline{underlined} are second.}
            \label{tab:CUB-CARS}
            \begin{tabular}{|l|cccc|cccc|}\hline
                & \multicolumn{4}{c|}{CUB-200-2011} & \multicolumn{4}{c|}{Cars-196} \\\hline
                 r@ & 1 & 2 & 4 & 8 & 1 & 2 & 4 & 8 \\\hline
                \cite{Ustinova_2016_NIPS} &  52.8 & 64.4 & 74.7 & 83.9 & - & - & - & -\\
                \cite{Sohn_2016_NIPS} & 51.0 & 63.3 & 74.3 & 83.2 & 71.1 & 79.7 & 86.5 & 91.6 \\
                \cite{Yuan_2017_ICCV} & 53.6 & 65.7 & 77.0 & 85.6 & 73.7 & 83.2 & 89.5 & 93.8 \\
                \cite{Opitz_2017_ICCV} & 55.3 & 67.2 & 76.9 & 85.1 & 78.0 & 85.8 & 91.1 & 95.1 \\
                \cite{Ge_2018_ECCV} & 57.1 & 68.8 & 78.7 & 86.5 & \underline{81.4} & \underline{88.0} & \underline{92.7} & \underline{95.7} \\\hline
                \ourlayer-32 & \textbf{60.1} & \textbf{72.1} & \textbf{81.7} & \textbf{88.3} & \textbf{82.6} & \textbf{89.2} & \textbf{93.5} & \textbf{96.0}\\
                \ourlayer-32-8 & \underline{58.1} & \underline{70.4} & \underline{80.3} & \underline{87.6} & 74.2 & 83.4 & 89.7 & 93.9 \\\hline
            \end{tabular}
        \end{center}
    \vspace{-1em}
    \end{table}

\section{Conclusion}\label{sec:ccl}
    In this paper, we explore codebook based second order representations that are intractable in practice.
    We propose a two-step factorization and a low-rank approximation designed to keep the richness of the second-order representation but with the compactness of the first-order.
    We provide ablation studies to confirm the necessity of a codebook pooling strategy, the impact of the different factorizations and the benefit of the low-rank approximation to control the computation cost.
    This representation named \ourlayer outperforms state-of-the-art methods on three image retrieval benchmarks and its low-rank approximation is state-of-the-art on two of them.




\bibliographystyle{IEEEbib}
\bibliography{biblio.bib}

\end{document}